\documentclass[10pt,twocolumn,letterpaper]{article}

\usepackage[applications]{wacv}      

\usepackage{graphicx}
\usepackage{amsmath}
\usepackage{amssymb}
\usepackage{booktabs}

\usepackage{adjustbox}
\usepackage{multirow}
\usepackage{xcolor}  
\usepackage{colortbl}
\usepackage{xcolor}
\usepackage{enumitem}
\usepackage{xspace}

\usepackage{makecell}
\usepackage{lipsum} 
\usepackage{gensymb} 
\usepackage{graphicx}
\usepackage{amssymb}
\usepackage{amsmath}
\usepackage{wrapfig}
\usepackage[linesnumbered,ruled,vlined]{algorithm2e}
\usepackage[mathscr]{euscript}
\usepackage{subcaption}
\usepackage{placeins}
\usepackage{stfloats}

\definecolor{LightCyan}{rgb}{0.88,1,1}
\definecolor{mygray}{gray}{0.9}
\usepackage[pagebackref,breaklinks,colorlinks]{hyperref}

\usepackage[capitalize]{cleveref}
\crefname{section}{Sec.}{Secs.}
\Crefname{section}{Section}{Sections}
\Crefname{table}{Table}{Tables}
\crefname{table}{Tab.}{Tabs.}

\def\wacvPaperID{1995} 
\def\confName{WACV}
\def\confYear{2025}

\begin{document}

\title{Continual Gesture Learning without Data via
Synthetic Feature Sampling}

\author{Zhenyu Lu\\
Carnegie Mellon University\\
{\tt\small zhenyulu@andrew.cmu.edu}
\and
Hao Tang\\
Carnegie Mellon University\\
{\tt\small hao.tang@vision.ee.ethz.ch}
}
\maketitle

\begin{abstract}
Data-Free Class Incremental Learning (DFCIL) aims to enable models to continuously learn new classes while retaining knowledge of old classes, even when the training data for old classes is unavailable. Although explored primarily with image datasets by researchers, this study focuses on investigating DFCIL for skeleton-based gesture classification due to its significant real-world implications, particularly considering gestures serve as one of the primary means of control and interaction in VR/AR, smart home systems and beyond.
In this work, we made an intriguing observation: skeleton models trained with base classes(even very limited) demonstrate strong generalization capabilities to unseen classes without requiring additional training. Building on this insight, we developed a new model-updating paradigm - \textbf{S}ynthetic \textbf{F}eature \textbf{R}eplay (\textbf{SFR}) that directly synthesize fake features within the embedding space from saved class prototypes, instead of generating synthetic data, to replay for old classes and augment for new classes (under a few-shot setting). Our proposed method showcases significant advancements over the state-of-the-art, achieving up to 15\% enhancements in mean accuracy across all steps and largely mitigating the accuracy imbalance between base classes and new classes.
\end{abstract}

\vspace{-3ex}
\section{Introduction}
\label{sec:intro}
\begin{figure}[htbp!]
\centering
\includegraphics[width=0.48\textwidth]{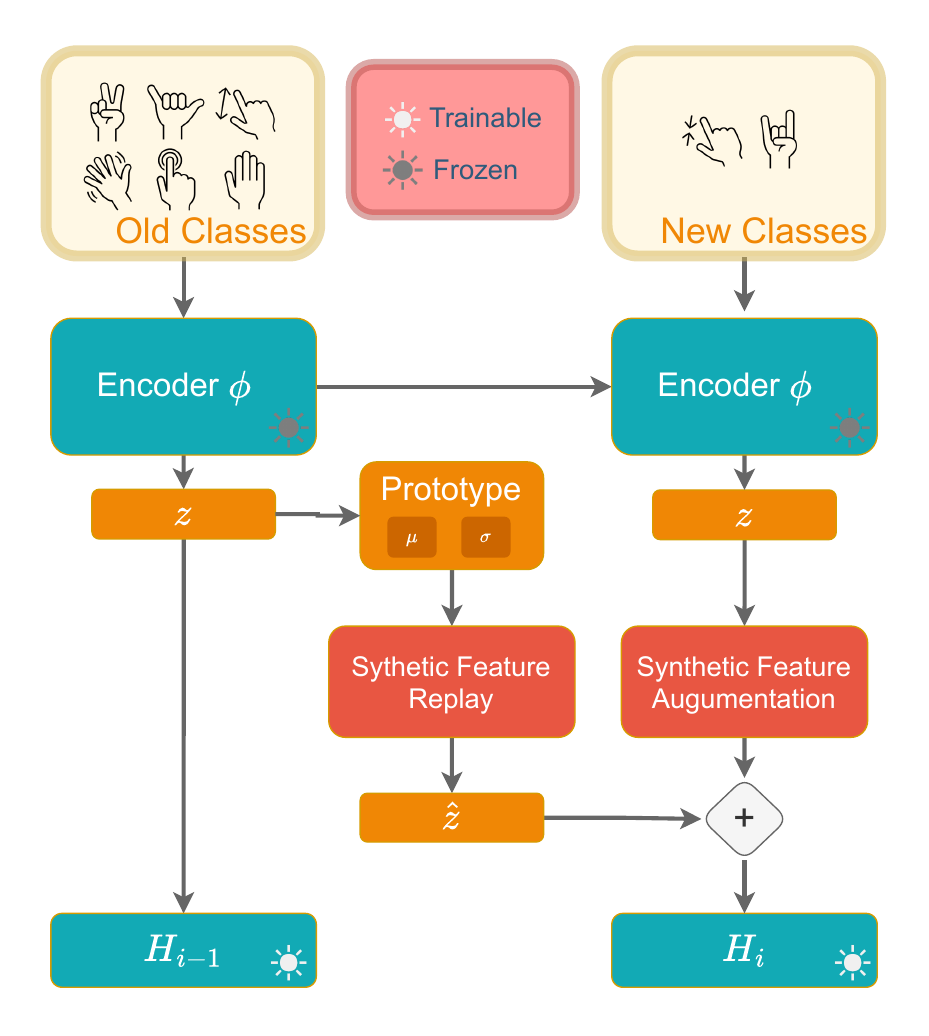}
\vspace{-3ex}\caption{Overall Pipeline. (1) Compute and save class prototypes from the embedding space. (2) Sample synthetic old class features from the saved class prototype using Alg.~\ref{algo:Replay}. (3) \textbf{Only in the few shot setting}, sample synthetic new class feature using Alg.~\ref{algo:Augmentation} to augment the new class data. (4) Combine new and old class features together to train the classifier.}
  \label{fig:pipeline}
  \vspace{-3ex}
\end{figure}

Class incremental learning (CIL) refers to the learning problem wherein models are continuously updated as new class data is introduced. This area has gained significant attention as researchers recognize that data in real-world scenarios typically arrives incrementally rather than all at once. Although considerable progress has been made in class incremental learning\cite{yang2019adaptive,lwf,icarl,gan_memory,fscil}, most of the research has focused on experimenting with images as the primary modality. However, the recent emergence of large pre-trained, visual or visual-language models, such as CLIP\cite{clip}, DINO\cite{dino}, SAM\cite{sam} and their variations, has demonstrated remarkable generalizability,  showcasing impressive zero-shot performance. This development, to some extent, challenges the significance of class incremental learning research focused solely on images. 

Therefore, we argue that class incremental learning research should address specific real-world problems or be differentiated by modalities, as different modalities may exhibit distinct interclass properties as the CIL algorithm primarily experimented on image datasets might not be a universal solution adaptable to other modalities. Unlike the abundance of image sources readily available from the Internet, human-related data sources are more limited due to factors such as labor costs and privacy concerns associated with collecting data from individuals. These types of data sources may align better with the continuous availability of data. 

Gesture-based interaction is the primary means of controlling AR/VR headsets \cite{megatrack, umetrack} and is also widely applied in various real-world scenarios, including smart homes, education, and more.  Class incremental learning for skeleton-based gesture recognition shows promising value; for example, users may register and customize their own gestures for various functions. However, data privacy and security are significant concerns for human-related computer vision tasks, as even skeletal gestures can inadvertently reveal personal identity information. Thus, restricting the setting to be data-free, meaning that only data used to train newly introduced classes is available without access to training data for old classes, becomes especially important and natural.

In this study, we explore Data Free Class Incremental Learning(DFCIL) for Skeleton Gesture, a new research domain introduced by BOAT-MI\cite{boat-mi}. Unlike BOAT-MI which involves data synthesis via model inversion and retraining of the whole model, we introduce a new model-updating paradigm called \textbf{S}ynthetic \textbf{F}eature \textbf{R}eplay (\textbf{SFR}). SFR confines the synthesis process to the embedding domain by modeling the class feature space as a multivariate normal distribution defined by a prototype (mean and covariance). This approach allows us to directly sample synthetic features to recap old class knowledge and facilitate the learning of new classes.

In summary, the key contributions of this paper are as follows:  
  \vspace{-1ex}
\begin{itemize}
    \item We reveal that models trained for skeleton gestures inherently possess better generalization ability for unseen classes without requiring additional training.
    \item Based on this observation, our proposed \textbf{SFR} circumvents the need for challenging synthetic data generation, which is typically used in existing DFCIL methods. Our approach is not only faster and simpler, but also significantly outperforms current SOTA methods across all tested datasets.
    \item Additionally, we examine the effect of shot size within the Few Shot Class Incremental Learning (FSCIL) context and found that sampling synthetic features for new classes combined with a carefully designed filtering mechanism can alleviate performance degradation caused by limited data availability.
\end{itemize}

\section{Related Work}
\label{sec:related_works}
\paragraph{Data Free Class Incremental Learning.}
Regularization-based approaches\cite{lwf,yang2019adaptive, kirkpatrick2017overcoming} are a major category to address catastrophic forgetting of old classes \cite{catastrophic} present in DFCIL. Knowledge distillation is a quite useful technique that is usually used in conjunction with other approaches.  \cite{lwf,icarl,wu2019large,xiao2014error,hou2019learning,castro2018end,lee2019overcoming } use knowledge distillation to ensure that the model is updated without deviating too much from its frozen copy of its previous state.  Model
expansion \cite{beef,dynamically,foster} is another effective technique, allowing the model to dynamically grow as needed. Replay-based methods for DFCIL usually involve generating pseudo-samples of previously seen classes to mitigate the forgetting of old class knowledge. Training a sample generator during the pretraining stage is an effective way \cite{gan_memory,deepgenerative_replay}, but model inversion is a technique that can invert the model to obtain fake old class samples without training an additional generator\cite{deepinversion,r-dfcil,abd}.

\paragraph{Few-Shot Class Incremental Learning.}
FSCIL \cite{fscil} is a setup that imposes additional restrictions on the sample size of incremental classes beyond those in DFCIL. One strategy widely used in few-shot class incremental learning is to keep the backbone encoder frozen and update only the classifier to adapt to new classes. These methods focus either on creating better pre-trained base models \cite{forward_compatible,cl_incremental,zhou2022few}, or designing novel classifiers \cite{evolved_classifiers}. Constructing a class prototype \cite{prototype_rect, gidaris2018dynamic,akyurek2021subspace,kukleva2021generalized}, originally proposed by ProtoNet \cite{prototypical}, primarily used for a few-shot learning, has proved useful in creating the classifier for FSCIL as well \cite{prototype_refinement, prototype_calibration}. Our approach also adopts such a frozen encoder strategy, but instead of directly using class prototypes to build the classifier, we sample synthetic features from class prototypes and train the classifier along with the sampled features. 

\paragraph{DFCIL for Skeleton-Based Gesture Recognition.} 
Using skeletal data, instead of raw image frames, for gesture recognition shows great value for its reduced data complexity and therefore effective processing and privacy protection. Deep learning approaches such as CNN-LSTM \cite{skeleton-convolutional}, GCNs\cite{st-gcn,infogcn,hierarchicallygcn}, transformers \cite{dg-sta,decoupled, stst, mformer}, typically show superior performance for skeleton-based gesture recognition. ST-GCN\cite{st-gcn} is a popular GCN-based model that can automatically learn both spatial and temporal patterns. DG-STA \cite{dg-sta} is a transformer-based model that applies multi-head attention to spatial and temporal edges, respectively, for hand gesture recognition.  


Class incremental learning for skeleton data, especially in the data-free setting, is still a valuable but underexplored domain. To the best of our knowledge, BOAT-MI \cite{boat-mi} represents pioneering work in investigating data-free class incremental learning of gestures by developing a model inversion-based approach to generate synthetic data and replay them during training new classes. Therefore, we follow their setup to run our experiments and benchmark against them.

\section{Preliminaries}
\label{sec:Preliminaries}
\paragraph{Problem Definition.}%
Data free class-incremental learning is the setup such that a model need to continue learning novel classes while not forgetting knowledge about old classes. Firstly, the entire dataset is divided into $\mathcal{N}$ subsets $\mathcal{D}_i = \{(X,Y) : Y \in \mathcal{C}_i, i \in T\} $. Here, $i$ denotes the $i$-th incremental session and $\mathcal{C}_i$ denotes the label space, noting that the class space introduced are exclusive between different steps. The base model is trained with $\mathcal{D}_0$ and the model needs to update based on the information from $\{\mathcal{D}_1,\mathcal{D}_2, ... ,\mathcal{D}_{\mathcal{N}-1}\}$ continuously, with data from previous incremental sessions no longer available. However, the test set for the $i$-th incremental session should encompass all classes seen before, such that $\mathcal{C}_i^{test} = \cup_{j=0}^{i} \mathcal{C}_j$. 

In addition to DFCIL, few-shot class-incremental learning (FSCIL) imposes a even stricter constraint where the sample size for each class during the incremental session is limited. Each task can be represented as an $N$-way $K$-shot classification task, where $N$ classes are learned in the current step, with $K$ samples available for each class.

\paragraph {Metric Learning.}
Metric learning or frozen encoder-based approaches are usually composed of two parts: a feature encoder $\phi$ and a classifier $\mathcal{H}$. Given the input and label of sample $(x,y)$, the feature of $x$ can be denoted as $z = \phi (x)$ and the predicted label can be obtained by $\hat y  = \mathcal{H}(z)$. 

The encoder $\phi$ pre-trained at the base stage is kept frozen for all upcoming incremental steps while only the classifier $\mathcal{H}_i$  for each step updates continuously. 

\paragraph{Instantaneous Forgetting Measure (IFM).}
The central dilemma in class incremental learning revolves around balancing the preservation of knowledge from old tasks with the ongoing learning of new tasks. However, many existing studies\cite{liu2022few,evolved_classifiers,zhou2022few,gan_memory} tend to over-prioritize overall accuracy as the primary evaluation metric, often overlooking the imbalance between the accuracy of the base classes and the incremental classes. 

Instantaneous Forgetting Measure\cite{dfcil_gesture} is an important metric evaluating such imbalances:
\begin{equation}
IFM = \frac{|L- G|}{ L + G } \times 100 \%,
\end{equation}
where $L$ represents the local accuracy solely over the novel classes introduced during the current step, while $G$ signifies the global accuracy over all classes encountered up to that stage.


\section{What to Replay? Embeddings or Data}%

Synthetic replay-based DFCIL approaches\cite{gan_memory, deepgenerative_replay, deepinversion, r-dfcil,boat-mi} typically involve training an additional generator or using model inversion to create fake data. However, generating those fake data poses significant challenges, as pre-training a generator during the previous step can be cumbersome, and the quality of data generated through model inversion may suffer, especially for skeletal gesture movements that entail both spatial and temporal changes. 

Thus, is there a way to bypass the challenging data synthesis step? 
Instead of traditional replay based DFCIL approaches which focus on data synthesis, we proposed a simple but effective paradigm that \textit{directly synthesize fake features from the embedding space} to replay the old class and facilitate the training of the new class, when the samples of the newly class are limited, as illustrated in Figure \ref{fig:pipeline}. The effectiveness of our approach is rooted in a key observation.



\begin{figure}[!]
    \centering
    \includegraphics[width=0.48\textwidth]{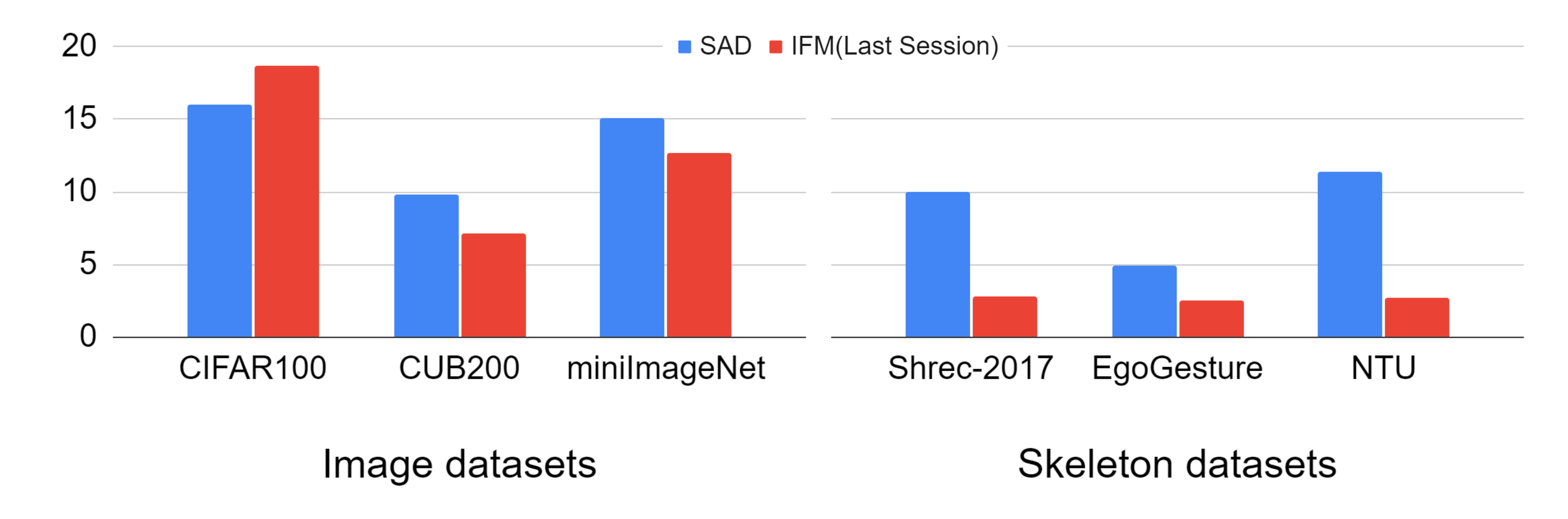}
        \centering
    \caption{A overall comparison between DFCIL on image datasets (Left) and skeleton datasets (Right), shows the performance on skeleton datasets generally superior to that on image datasets (We use our proposed approach to generate experiment result for skeleton datasets and use TEEN \cite{prototype_calibration} for image datasets as it is the SOTA method for image datasets). Session Accuracy Delta (SAD) measures the gap between the accuracy of the base session and the accuracy of the final session after the model has processed all data. IFM (last session) measures the Instantaneous Forgetting Measure of the last session. For both metrics, lower is better.}
    \label{fig: compare_image_skeleton}
    \vspace{-3ex}

\end{figure}
\begin{figure*}[!h]

    \centering
    \begin{subfigure}[b]{0.45\textwidth}
        \centering
        \includegraphics[width=\textwidth]{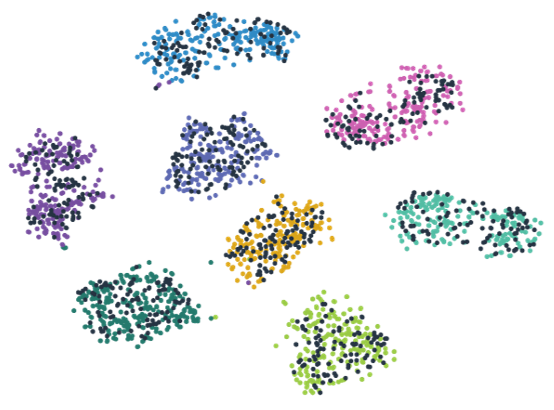}
         \caption{T-SNE visualization comparing ground truth features and sampled synthetic features. The sampled features are colored in black.}
        \label{fig:feature_sampling}
    \end{subfigure}
    \hfill
    \begin{subfigure}[b]{0.48\textwidth}
        \centering
        \includegraphics[width=\textwidth]{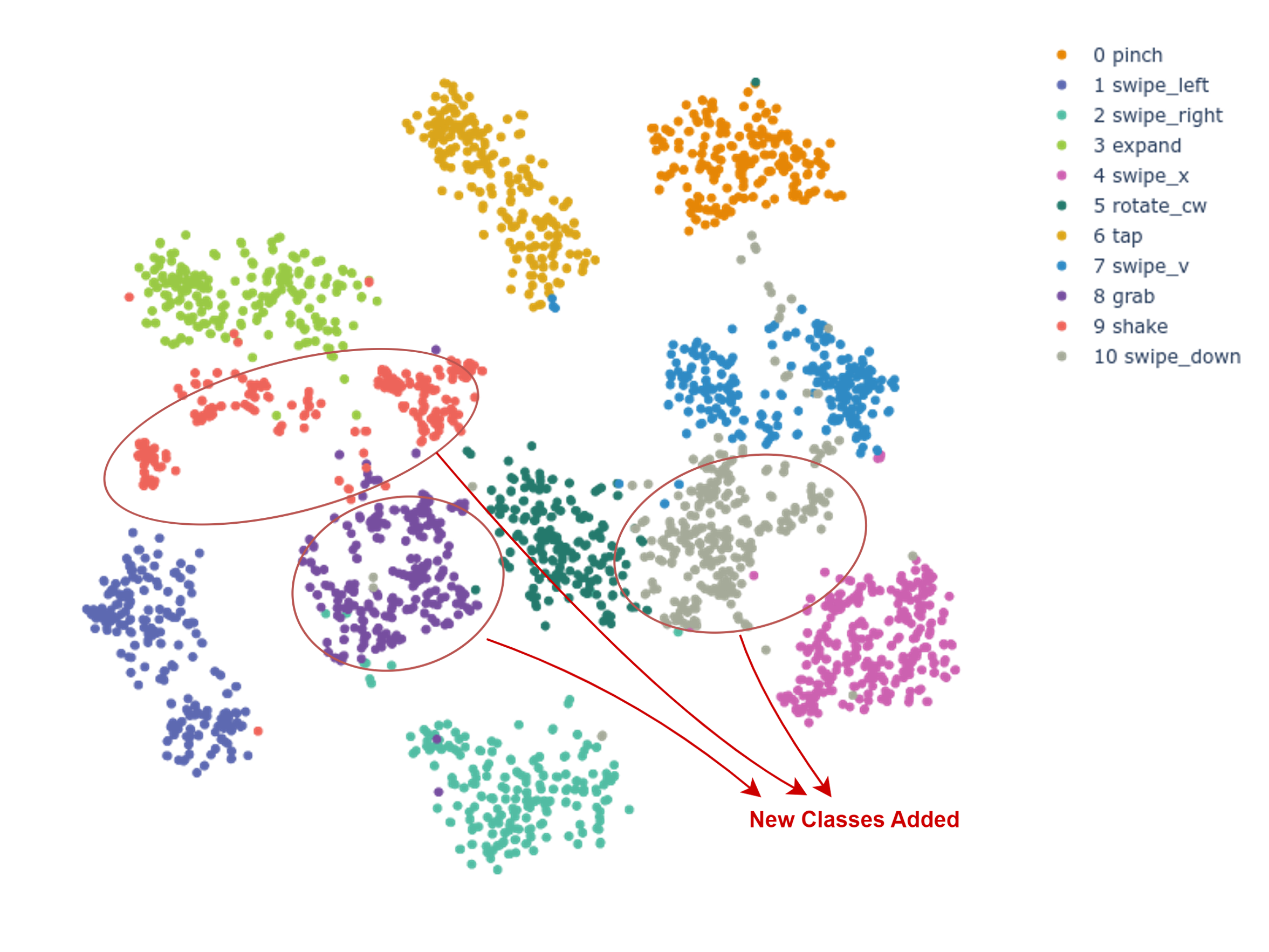}
        \caption{T-SNE visualization comparing features of old and new classes. The features of the newly introduced classes are circled in red.}
        \label{fig:increm_viz}
    \end{subfigure}
    \caption{A toy experiment on Shrec-2017, three new hand gesture classes were added to a model pre-trained on eight base classes. A similar experiment was conducted on an image dataset for comparison, and the details can be found in the appendix.}

\end{figure*}

\paragraph{Skeleton Models Trained on Base Classes Can Generalize to Unseen Classes Well.}%
Frozen encoder-based approaches are widely used in FSCIL. However, when applied to image domains, they often suffer from severe accuracy imbalance, and the model struggles on new classes, as illustrated in Figure \ref{fig: compare_image_skeleton}, which explains why this strategy is not commonly used for DFCIL when the sample size is abundant. As noted in TEEN \cite{prototype_calibration}, new classes tend to be classified into the most similar base classes. Despite their attempts to address this issue through calibration, the accuracy gap between base classes (70\%-80\%) and new classes (40\%-50\%) remains substantial. This discrepancy leads to an Incremental Forgetting Measure (IFM) score of approximately 20\%, indicating the model's difficulty in adapting to unseen classes. 

However, we found that using a frozen encoder is particularly well suited for skeleton modality, as our experiments unveiled an interesting discovery: models trained on skeletal-based gesture movement data can automatically generalize well to new, unseen classes without additional training. This enables the formation of compact feature spaces that establish distinguishable boundaries from old and new classes. A simple experiment on Shrec-2017\cite{shrec}, as shown in Figure \ref{fig:increm_viz}, illustrates how within-class features cluster together, forming compact class feature spaces with minimal overlap between different classes. This phenomenon appears to be specific to skeletal data (or potentially other structured modalities). Taking a hand gesture as an example, regardless of the specific movement (e.g., pinch or swipe), all data samples are fundamentally hand motions, and the structure should always adhere to the physical appearance of a hand. Thus, treating all data as a single class with various subclasses, explainable by semantic relationships, seems reasonable. Previous studies \cite{norouzi2013zero,farhadi2009describing,lampert2009learning,frome2013devise,palatucci2009zero,mikolov2013efficient} have shown that both large image and language models demonstrated a strong relationship between feature space and semantic information, especially within a large category. Furthermore, an observation that supports this claim is the significantly higher DFCIL performance on CUB200 (lower SAD and IFM in the last session) compared to two other image datasets, as shown in Figure \ref{fig: compare_image_skeleton}. This can possibly be attributed to the fine-grain nature of the CUB200 which contains 200 subcategories of a single large class - bird. The skeletal encoder, even trained with very limited data, appears to be capable of capturing such a feature relationship. This observation forms the basis for our feature sampling approach. 
    \vspace{-3ex}
\paragraph{Synthetic Feature Sampling.}%
First, the model is trained with initial subset of dataset $\mathcal{D}_0$. Then, the model's feature encoder $\phi$ is kept frozen for all incremental sessions and only the classifier $\mathcal{H}$ is continuously updated. Prior to training the current classifier $\mathcal{H}_i$ , only the class prototypes for all seen classes before the current incremental step are saved and kept for future use as well.

The class prototypes can be described by the mean $\mu_k$ and covariance $\Sigma_k$, to provide an accurate approximation of class feature space. Then, we can directly synthesize embeddings from the prototypes by making the assumption that each class feature space follows a multivariate Gaussian distribution (we further investigate the correctness and feasibility of the Gaussian assumption in the appendix). The sampling processing can be formulated as:
\begin{equation}
\label{sampling}
(\hat Z_k,Y_k) | \hat Z_k \sim \mathcal{N} (\mu_{k}, \Sigma_k), \forall  Y_k \in \mathcal{C},
\end{equation} 
where $\hat Z_k$ is the synthetic feature space of class $Y_k$ sampled from the distribution, $C$  is the label space.


\begin{algorithm}[!t]
  \KwIn{Training  features from  previous session $ S_{i-1} = (Z_k, Y_k) : Y_k \in \mathcal{C}_{i-1}$, pre-trained classifier $\mathcal{H}_{i-1}$}

  \For{$Y_k \in \mathcal{C}_{i-1}$ }{ 
    Filtered features $Z_k \newline = \{ z_k \mid z_k \in Z_k, \; y_k = \mathcal{H}_{i-1}(z_k) \}$;
\textcolor{gray}{\#reject features if $y_k \neq \mathcal{H}_{i-1}(z_k))$  }\;
    Compute the mean $\mu_{k} = \mathbb{E}[Z_{k}]$ and the covariance $\Sigma_k = \mathbb{E}[Z_{k} - \mu_k)(Z_{k} - \mu_k)^{T}]$ \;
    [Optional] \If{$\Sigma_k$ is not PSD }{
    $Z_k,  u_k = SVD(Z_k)$ \text{\textcolor{gray}{\#reduce dimension }} \;
    Recompute the mean and the covariance on reduced dimension space
    }   
    Sample synthetic features $\hat{Z}_k$ from $\mu_{k}$ and $\Sigma_{k}$ using Eq. \eqref{sampling} \;
    [Optional] \If{$\Sigma_k$ is reduced}{
    $\hat{Z}_k  = \hat{Z}_k * u_k$ \text{\textcolor{gray}{\# revert it back}} \;
    }
    Filtered synthetic features $\hat{Z}_k \newline = \{ \hat{z}_k \mid \hat{z}_k \in \hat{Z}_k, \; y_k = \mathcal{H}_{i-1}(\hat{z}_k) \}$ \
    \textcolor{gray}{\#reject features if $y_k \neq \mathcal{H}_{i-1}(\hat{z}_k))$  }\;
  }

 \caption{Synthetic Feature Replay}
  \label{algo:Replay}
 
\end{algorithm}

Considering that feature representations often reside in high-dimensional spaces, a potential issue arises with the dimensionality-collapse problem. This arises when the computed class covariance fails to be positive semi-definite, contradicting the assumption of a normal distribution. One proposed solution is to employ singular value decomposition (SVD) to reduce the dimensionality of the feature space. Subsequently, sampling can be conducted in this reduced feature space, followed by reverting the sampled features back to their original dimension size, as elaborated in Alg. \ref{algo:Replay}.

After the sampling process is complete,  we can obtain the real features $ Z$  of classes newly introduced in the current incremental session by simply feeding the data into the frozen encoder $\phi$, along with the fake features $\hat Z$ sampled from saved old class prototypes. These real and fake features are then combined to train the classifier $\mathcal{H}_i$.  


As shown in Figure \ref{fig:feature_sampling}, it is evident that there is a nearly complete overlap, suggesting that the sampled features accurately mirror the feature space of the ground truth class. Thus, it comes naturally that the classifier trained with such good features can achieve excellent performance. 


\subsection{Effect of Supervised Contrastive Learning}%
Model trained with constraive learning are proven to generate more robust and generailzable representation, making it effective for multiple downstream tasks, and have a superior transfer ability than its cross-entropy counterpart \cite{cl_transferability}. Some previous work has shown its effectiveness in continual learning studies \cite{cl_continual,cl_incremental}.


Tables \ref{tab:shrec-2017} and \ref{tab:ego_gesture} show the performance when the base model is trained with supervised contrastive loss (Supcon) and cross entropy. 
For Shrec-2017, which is a small dataset, the model trained with Supcon indicates a great performance for both overall accuracy and more significantly for the IFM score, which is half compared to model trained with cross entropy. However, for the much larger Ego-Gesture3D, the significance of contrastive learning is not significant.

\subsection{Effect of Replay Buffer Size}%
\begin{figure}[!]
\centering
\includegraphics[width=0.48\textwidth]{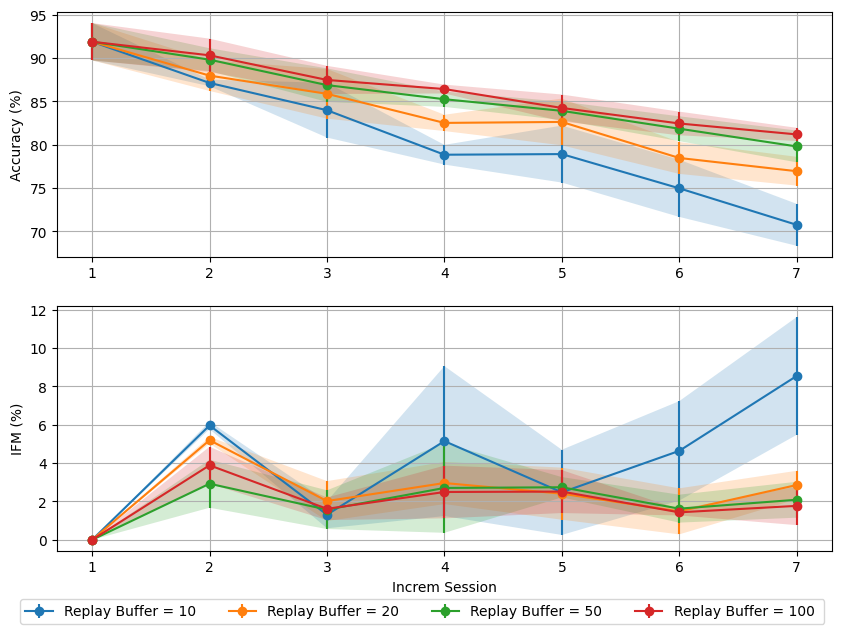}
\vspace{-3ex}\caption{The impact of replay buffer size on performance when evaluating on Shrec-2017. Detailed results can be found in appendix.}
  \label{fig:replay_size_effect}
\vspace{-4ex}
\end{figure}

\begin{table*}[!htp]
\small
\centering
\caption{Results (average of 3 runs with different class order) for data free class-incremental learning over six tasks in Shrec-2017. SFR-CE indicates that the base model is trained using cross-entropy, while SFR-CL indicates that the base model is trained using supervised contrastive loss.
\vspace{-1ex}}
\begin{adjustbox}{width=2.1\columnwidth,center}
\begin{tabular}{lccccccccccccccc}
\toprule
            Method &    Task 0 &    \multicolumn{2}{c}{Task 1} &    \multicolumn{2}{c}{Task 2} &    \multicolumn{2}{c}{Task 3} &    \multicolumn{2}{c}{Task 4} &    \multicolumn{2}{c}{Task 5} &    \multicolumn{2}{c}{Task 6} & \multicolumn{2}{c}{Mean (Task $1 \rightarrow 6)$} \\
\midrule
            Oracle &      \multicolumn{14}{c}{89.4} \\
\midrule
        
          &       & G$\uparrow$ & IFM $\downarrow$ & G$\uparrow$ & IFM$\downarrow$ & G$\uparrow$ & IFM $\downarrow$  & G$\uparrow$ & IFM $\downarrow$  & G$\uparrow$ & IFM $\downarrow$  & G$\uparrow$ & IFM$\downarrow$ & G$\uparrow$ & IFM$\downarrow$ \\ 
          \hline
    Base \cite{lwf} &  90.5 & 78.3 & 7.8 & 53.3 & 29.6 & 30.0 & 53.7 & 27.9 & 56.0 & 12.2 & 78.3 & 11.9 & 78.6 & 35.6 & 50.7 \\
    Fine tuning \cite{lwf} &  90.5   & 78.6 & 6.0 & 57.4 & 26.2 & 34.2 & 48.8 & 34.0 & 48.6 & 16.4 & 71.8 & 16.1 & 71.9 & 39.5 & 45.5 \\
    Feature extraction \cite{lwf} &  90.5  & 79.9 & 9.3 & 69.3 & 12.6 & 62.3 & 19.1 & 56.5 & 24.1 & 50.6 & 24.3  & 45.1 & 32.4 & 60.6 & 20.3 \\
    LwF \cite{lwf} &    90.5 & 79.8 & 9.3 & 64.1 & 20.2 & 31.9 & 50.2 & 29.1 & 53.1 & 13.1 & 76.8 & 11.5 & 79.3 & 38.2 & 48.2 \\
    LwF.MC \cite{abd} &  90.5    & 56.0 & 10.9 & 32.6 & 39.5 & 21.8 & 58.5 & 19.3 & 61.6 & 16.7 & 53.3  & 16.1 & 45.4 & 27.1 & 44.9 \\
    DeepInversion \cite{deepinversion} &  90.5    & 79.9 & 4.3 & 65.9 & 14.9 & 53.1 & 29.5 & 49.5 & 32.8 & 34.2 & 47.7  & 32.1 & 49.7 & 52.5 & 29.8  \\
    ABD \cite{abd} &   90.5   & 78.8 & 4.0 & 64.6 & 12.6 & 54.3 & 20.3 & 53.2 & 24.6 & 46.1 & 20.0  & 40.4 & 23.3 & 56.2 & 17.5 \\
    R-DFCIL \cite{r-dfcil} &  90.5  & 78.7 & 3.3 & 65.5 & 4.5 & 54.4 & 20.5 & 49.8 & 26.9 & 41.5 & 25.0 & 38.6 & 33.3 & 54.8 &  18.9 \\
    BOAT-MI \cite{boat-mi} & 90.5  & 83.7 & 4.5 & 76.0 & 7.3 &  71.4 & 7.0 &  69.4 & 13.3 &  64.1 & 9.5 &  58.1 & 11.2 & 70.5 & 8.8 \\
    TEEN \cite{prototype_calibration}& 91.2 & 87.7 & 12.7 & 80.6 & 27.7 & 76.3 & 28.6 & 72.9 & 27.4 & 68.4 & 29.4 & 64.7 & 31.1 & 77.4 & 22.4 \\
    \midrule 
    Our Approaches\\
    \midrule
    \rowcolor{mygray}
    SFR-CE & 91.1 & 89.6 & 2.1 & \textbf{87.8} & 3.4 & 85.1 & 2.7 & 83.8 & 2.2 & 81.6 & 3.8 & 79.8 & 5.0 & 85.5 & 2.8 \\
    \rowcolor{mygray}
    SFR-CL & \textbf{91.9} & \textbf{90.1} & \textbf{2.1} & 87.3 & \textbf{1.1} & \textbf{85.9} & \textbf{2.4} & \textbf{84.0} & \textbf{1.9} & \textbf{82.5} & \textbf{1.6} & \textbf{81.1} & \textbf{2.8} & \textbf{86.1} & \textbf{1.7}  \\

\bottomrule
\end{tabular}
\end{adjustbox}
\label{tab:shrec-2017}
\end{table*}

\begin{table*}[!htp]
\small
\centering
\vspace{-3mm}
\caption{Results  (average of 3 runs with different class order) for data free class-incremental learning over six task in Ego-Gesture3D. \vspace{-1ex}}
\begin{adjustbox}{width=2.1\columnwidth,center}
\begin{tabular}{lccccccccccccccc}
\toprule
            Method &    Task 0 &    \multicolumn{2}{c}{Task 1} &    \multicolumn{2}{c}{Task 2} &    \multicolumn{2}{c}{Task 3} &    \multicolumn{2}{c}{Task 4} &    \multicolumn{2}{c}{Task 5} &    \multicolumn{2}{c}{Task 6} & \multicolumn{2}{c}{Mean (Task $1 \rightarrow 6)$} \\
\midrule
            Oracle &      \multicolumn{14}{c}{75.8} \\
\midrule
        
          &       & G$\uparrow$ & IFM $\downarrow$ & G$\uparrow$ & IFM$\downarrow$ & G$\uparrow$ & IFM $\downarrow$  & G$\uparrow$ & IFM $\downarrow$  & G$\uparrow$ & IFM $\downarrow$  & G$\uparrow$ & IFM$\downarrow$ & G$\uparrow$ & IFM$\downarrow$ \\ 
          \hline

    Base \cite{lwf} &   78.1 & 60.4 & 13.5 & 18.9 & 63.3 & 9.3 & 82.4  & 8.0 & 84.3  & 5.9 & 88.5  & 5.9 & 88.3 & 18.1 & 70.0 \\
    Fine tuning \cite{lwf} &  78.1     & 59.2 & 10.4 & 15.9 & 66.5 & 9.3 & 82.2  & 7.7 & 84.4  & 5.2 & 89.7  & 5.0 & 90.1 & 17.1 & 70.6 \\
    Feature extraction \cite{lwf} &  78.1      & 69.8 & 14.5 & 60.8 & 17.2 & 51.5 & 30.4 & 46.4 & 33.7 & 41.2 & 39.4 & 36.8 & 42.9 & 51.1 & 29.7 \\
    LwF \cite{lwf} &   78.1    & 69.1 & \textbf{0.1} & 43.0 & 27.3 & 18.8 & 67.3 & 11.3 & 78.7 &  6.5 & 87.5 &  6.3 & 87.7 & 25.8 & 58.1  \\
    LwF.MC \cite{abd} &  78.1     & 36.8 & 9.8 & 24.2 & 41.3 & 17.0 & 64.6 & 12.8 & 73.3 & 10.2 & 78.3 & 9.7 & 80.2 & 18.4 & 57.9 \\
    DeepInversion \cite{deepinversion} &   78.1    & 68.1 & 14.1 & 44.3 & 32.2 & 24.7 & 59.2 & 16.2 & 71.2 & 11.6 & 78.8 & 10.0 & 81.0 & 29.1 & 56.1 \\
    ABD  \cite{abd}  &    78.1   & 68.8 & 14.8 & 61.1 & 17.1 & 54.3 & 26.2 & 49.0 & 31.1 & 43.2 & 37.5 & 39.0 & 40.6 & 52.6 & 27.9 \\
    R-DFCIL \cite{r-dfcil} & 78.1 & 70.3 & 3.0 & 61.4 & 4.1 & 53.2 & 14.9 & 46.1 & 17.8 & 39.2 & 35.0 & 35.2 & 36.3 & 50.9 & 18.5 \\
    BOAT-MI \cite{boat-mi} & \textbf{78.1} & 75.3 & 4.6 & 70.9 & 7.3 & 64.0 & 15.1 & 58.6 & 16.3 & 52.3 & 26.8 & 45.8 & 32.2 & 61.2 & 17.1 \\
    TEEN \cite{prototype_calibration} & 77.8  & \textbf{76.2} & 12.8 & \textbf{74.6} & 13.3 & 73.1 & 11.9 & 72.2 & 10.9 & 70.7 & 11.4 & 69.4 & 10.9 & \textbf{73.4} & 10.2\\
    \midrule
    Our Approaches\\
    \midrule
    \rowcolor{mygray}
    SFR-CE & 75.7 & 75.5 & 5.7 & 74.3 & \textbf{2.3} & \textbf{73.1} & \textbf{2.2} & \textbf{72.3} & \textbf{2.4} & \textbf{71.3} & \textbf{1.9} & \textbf{70.4} & \textbf{2.5} & 73.2 & \textbf{2.4} \\
    \rowcolor{mygray}
    SFR-CL & 75.3 & 74.6 & 5.6 & 73.6 & 3.7 & 72.6 & 3.7 & 71.8 & 3.7 & 70.6 & 2.7 & 69.8 & 3.0 & 72.6 & 3.2 \\
\bottomrule \vspace{-5ex}
\end{tabular}
\end{adjustbox}
\label{tab:ego_gesture}
\end{table*}

\begin{figure*}[!htp]
    \centering
    \begin{subfigure}[b]{0.48\textwidth}
        \centering
        \includegraphics[width=\textwidth]{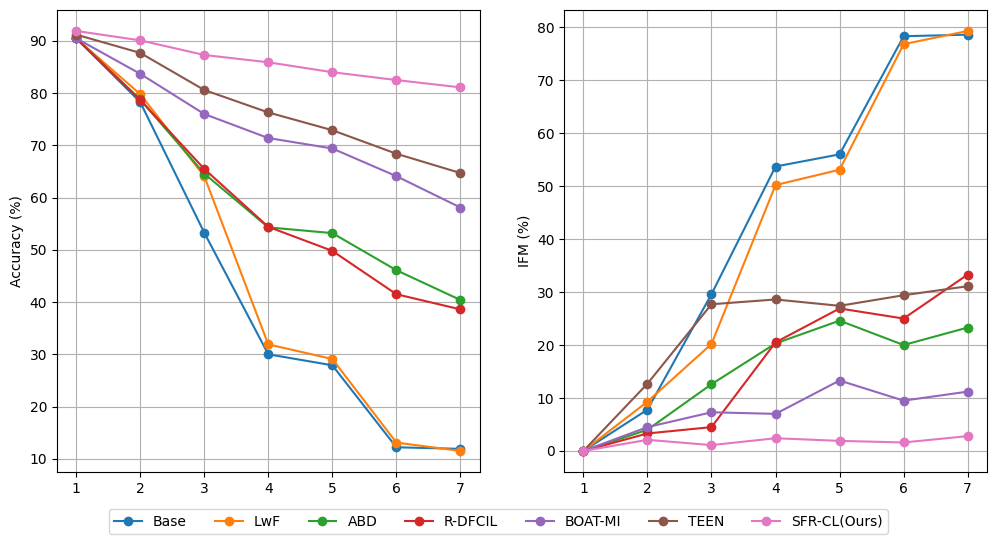}
         \caption{Shrec-2017}
        \label{fig:shrec_sota}
    \end{subfigure}
    \hfill
    \begin{subfigure}[b]{0.48\textwidth}
        \centering
        \includegraphics[width=\textwidth]{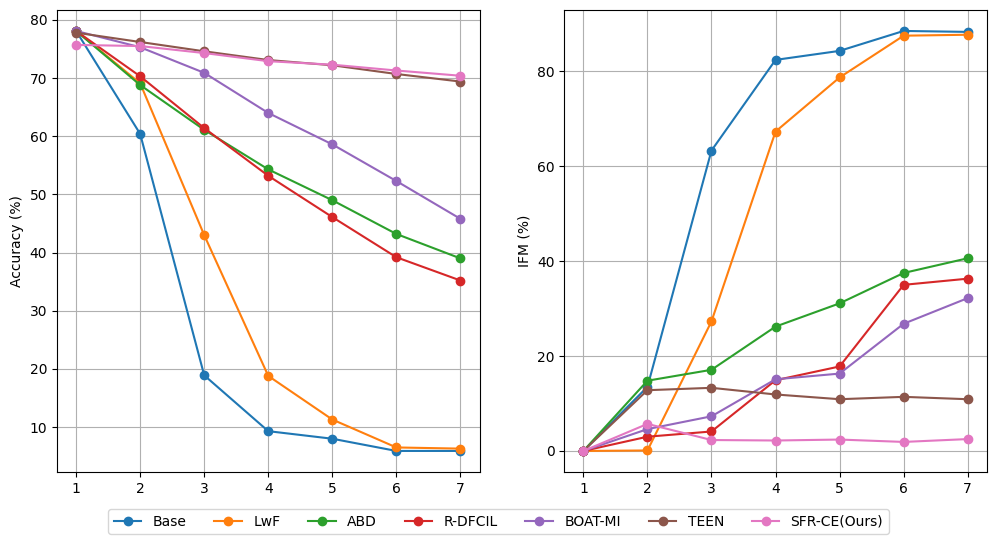}
        \caption{Ego-Gesture3D}
        \label{fig:ego_sota}
    \end{subfigure}
    \caption{SOTA comparison of global accuracy and IFM for all incremental sessions.}
 \label{fig:sota_comp}
 \vspace{-3ex}
\end{figure*}
We investigated how the size of synthetic features sampled affect the performance. As shown in Figure \ref{fig:replay_size_effect}, performance improves as the buffer size increases from 10 to 20, 50, and nearly saturates at 100. A limited replay buffer size also results in unstable outcomes across different trials, leading to an increased standard error. This highlights the necessity of class prototypes described by mean and covariance, allowing for the sampling of more diverse examples.

\subsection{SOTA Comparison}%
Although TEEN \cite{prototype_calibration} is primarily designed for FSCIL, we re-implemented this metric learning approach for our set-up based on our observation, found that it performs well and already surpasses the SOTA previously established by BOAT-MI \cite{boat-mi}. 
As shown in Figure \ref{fig:sota_comp}, our proposed approach further improves the score, especially by achieving more balanced performance (significantly lower IFM). In both data sets, the accuracy gap between two incremental sessions is impressively within 2\%. For the Shrec-2017 dataset, our method shows a 16\% improvement in global accuracy and a 7\% reduction in IFM measures compared to BOAT-MI. In the final and most challenging stage, our approach achieves 81\% accuracy — a 23\% improvement — while maintaining a very low IFM score, demonstrating an excellent accuracy balance between base and new classes. For the Ego-Gesture3D dataset, when comparing the final step accuracy after training on all classes to the oracle setup in which the model is trained on all classes simultaneously, there is only a 5\% decline, which is impressive among conventional DFCIL tasks.  Previous approaches testing on image datasets typically suffered from significant accuracy imbalances between new and old classes, indicated by high IFM scores that worsen with continued incremental learning. Our approach significantly mitigates this effect, keeping the IFM within 5\% for all incremental sessions in both tested datasets. 
    \vspace{-3ex}
\paragraph{Efficiency and Privacy.} 
In addition to its superior performance, feature synthesis offers the significant advantage of low computational cost. This approach only requires simple sampling from the embedding space, which typically takes just a few seconds or less. In contrast, other replay-based methods often involve training generators and creating pseudo-data, both of which are computationally expensive. For instance, BOAT-MI relies on the model inversion process, which can take hours just to generate fake samples. Furthermore, synthetic data might still reveal certain characteristics of the real training samples, which may conflict with the goal of data-free methods aimed at enhancing privacy. SFR also safeguards privacy by confining the synthesis process within the embedding domains.

\begin{table*}[!]
\small
\centering
\caption{Results  (average of 3 runs with different shots selected) for few shot class-incremental learning in Shrec-2017.  \vspace{-1ex}}
\begin{adjustbox}{width=2\columnwidth,center}
\begin{tabular}{lccccccccccccccc}
\toprule
            Method &    Task 0 &    \multicolumn{2}{c}{Task 1} &    \multicolumn{2}{c}{Task 2} &    \multicolumn{2}{c}{Task 3} &    \multicolumn{2}{c}{Task 4} &    \multicolumn{2}{c}{Task 5} &    \multicolumn{2}{c}{Task 6} & \multicolumn{2}{c}{Mean (Task $1 \rightarrow 6)$} \\

\midrule
        
          &   & G$\uparrow$ & IFM $\downarrow$ & G$\uparrow$ & IFM$\downarrow$ & G$\uparrow$ & IFM $\downarrow$  & G$\uparrow$ & IFM $\downarrow$  & G$\uparrow$ & IFM $\downarrow$  & G$\uparrow$ & IFM$\downarrow$ & G$\uparrow$ & IFM$\downarrow$ \\ 
          \hline
        
     \textbf{BOAT-MI} (\textit{Full-Shot})  & 78.1 & 75.3 & 4.6 & 70.9 & 7.3 & 64.0 & 15.1 & 58.6 & 16.3 & 52.3 & 26.8 & 45.8 & 32.2 & 61.2 & 17.1 \\
    \textbf{TEEN} (\textit{5-Shot}) & 88.7  & 83.4 & 30.5 & 76.1 & 40.6 & 71.3 & 40.8 & 72.4 & 23.4 & 68.0 & 25.5 & 63.8 & 27.9 & 74.8 & 27.0 \\
    \textbf{SFR} (\textit{Full-Shot})  & 89.4 & 88.4 & 3.3 & 87.3 & 2.1 & 86.0 & 0.5 & 85.3 & 0.2 & 83.0 & 0.2 & 82.3 & 1.9 & 85.9 & 1.2\\
    \textbf{SFR} (\textit{10-Shot}) & 89.4 & 88.0 & 6.2 & 86.3 & 5.8 & 81.8 & 12.1 & 85.4 & 1.2 & 78.8 & 6.9 & 78.9 & 6.2 & 84.1 & 5.5 \\
    \midrule
    \rowcolor{mygray}
   \textbf{SFR} (\textit{5-Shot}) & 89.4 & 85.9 & 20.7 & 82.8 & 15.8 & 79.9 & 13.2 & 78.5 & 9.8 & 75.9 & 11.1 & 74.4 & 10.1 & 81.0 & 11.5\\
     \rowcolor{mygray}
   \textbf{SFR-A} (\textit{5-Shot})  & 89.4 & 87.2 & 10.0 & 84.2 & 10.5 & 79.2 & 9.7 & 79.8 & 5.9 & 75.0 & 10.2 & 74.4 & 5.9 & 81.3 & 7.5  \\

\bottomrule
\end{tabular}
\end{adjustbox}
\label{tab:shrec-2017_fs}
\end{table*}


\begin{table*}[!]
\small
\centering
\vspace{-3mm}
\caption{Results  (average of 3 runs with different shots selected) for few shot class-incremental learning in Ego-Gesture3D. \vspace{-1ex}}
\begin{adjustbox}{width=2\columnwidth,center}
\begin{tabular}{lccccccccccccccc}
\toprule
            Method &    Task 0 &    \multicolumn{2}{c}{Task 1} &    \multicolumn{2}{c}{Task 2} &    \multicolumn{2}{c}{Task 3} &    \multicolumn{2}{c}{Task 4} &    \multicolumn{2}{c}{Task 5} &    \multicolumn{2}{c}{Task 6} & \multicolumn{2}{c}{Mean (Task $1 \rightarrow 6)$} \\

\midrule
        
          &       & G$\uparrow$ & IFM $\downarrow$ & G$\uparrow$ & IFM$\downarrow$ & G$\uparrow$ & IFM $\downarrow$  & G$\uparrow$ & IFM $\downarrow$  & G$\uparrow$ & IFM $\downarrow$  & G$\uparrow$ & IFM$\downarrow$ & G$\uparrow$ & IFM$\downarrow$ \\ 
          \hline
    \textbf{BOAT-MI} (\textit{Full-Shot}) & 78.1 & 75.3 & 4.6 & 70.9 & 7.3 & 64.0 & 15.1 & 58.6 & 16.3 & 52.3 & 26.8 & 45.8 & 32.2 & 61.2 & 17.1 \\
    
    \textbf{TEEN} (\textit{5-Shot}) & 77.1  & 74.6 & 28.9 & 72.5 & 26.9 & 70.1 & 27.0 & 69.0 & 22.7 & 66.5 & 24.9 & 64.6 & 24.7 & 70.6 & 22.2 \\

   \textbf{SFR} (\textit{Full-Shot}) & 75.2 & 75.6 & 2.5 & 73.1 & 3.0 & 72.7 & 1.7 & 72.7 & 2.4 & 70.2 & 2.2 & 69.5 & 3.2 & 72.7 & 2.1 \\
   \textbf{SFR} (\textit{10-Shot}) & 75.2  & 73.6 & 14.5 & 71.8 & 9.0 & 68.9 & 11.5 & 66.5 & 8.1 & 66.3 & 6.5 & 63.0 & 13.6 & 69.4 & 9.0 \\

    \midrule
    \rowcolor{mygray}
   \textbf{SFR} (\textit{5-Shot}) & 75.2 & 72.1 & 15.8 & 68.4 & 25.2 & 65.1 & 18.6 & 64.4 & 15.2 & 60.8 & 19.3 & 58.9 & 20.2 & 66.4 & 16.3 \\
    \rowcolor{mygray}
    \textbf{SFR-A} (\textit{5-Shot}) & 75.2  & 72.3 & 10.8 & 69.1 & 14.2 & 66.6 & 15.7 & 66.4 & 12.5 & 62.6 & 11.9 & 58.3 & 18.0 & 67.2 & 11.9\\
    \bottomrule
\multicolumn{9}{l}{*\textbf{SFR-A} denotes feature augmentation is enabled while \textbf{SFR} is disabled. } \\
 \vspace{-5ex}
\end{tabular}
\end{adjustbox}
\label{tab:ego_gesture_fs}
\vspace{-3mm}
\end{table*}
\section{Few-Shot Class Incremental Learning}%
\label{sec:fewshot}

In the real world, the number of samples that the user can provide to update the model is usually limited. For instance, if the user would like to register a personalized gesture, it is crucial to balance the need for data with the user experience, as asking for too much data can be burdensome and discourage users from using the developed function. Therefore, while our primary focus is on DFCIL, we also examine our approach within FSCIL to evaluate its performance with limited data availability.
\subsection{Effect of Shot Size}%

To investigate the impact of sample size on the performance, we conducted experiments across three settings, 5-Shot, 10-Shot, and Full-Shot.  The comparison results are presented in Tables \ref{tab:shrec-2017_fs} and \ref{tab:ego_gesture_fs}. We can easily see that, for both datasets, having a larger sample size can significantly improve the performance, particularly for IFM. For Shrec-2017 dataset, employing a 5-Shot setup results in a roughly 9\% decrease in accuracy and yields a significantly higher IFM score, increasing from 2\% to 13.2\%, compared to the Full-Shot setup. This phenomenon contrasts with many other frozen encoder-based approaches that achieve SOTA performance on image datasets. For example, \cite{evolved_classifiers} shows marginal or even slightly worse performance with increased data availability.
\begin{figure}[tbp!]
\centering
\includegraphics[width=0.45\textwidth]{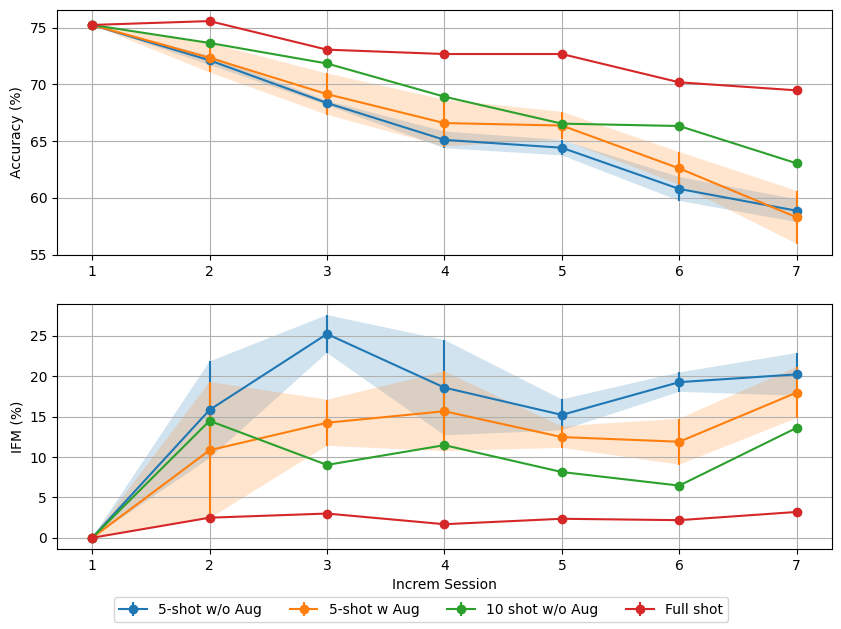}
\caption{The effect of shot size and augmentation on Ego-Gesture3D. Both the blue and orange lines indicate the 5-Shot setting with the orange lines denotes the synthetic feature augmentation enabled.  }
  \label{fig:few_shot}
  \vspace{-3ex}
\end{figure}

The positive relationship between shot size and performance can be explained by the fact that given more training data, there is a greater likelihood of uncovering the true and robust class distribution without being influenced by outliers or unrepresentative samples. Consequently, this allows for the generation of more diverse yet still within-distribution synthetic features. 

\subsection{Feature Augmentation}%

In light of the positive relationship between performance and the size of training samples, the instinctive approach is to augment or generate additional synthetic features for classifier training.
However, using less data to construct prototypes increases the likelihood of biasing the simulated class distribution away from the true distribution. Then, the problem becomes how can we sample the correct features from a likely biased distribution. In the feature replay method previously outlined, where the classifier trained for the previous step is available, candidate features can be inputted into the previous classifier, and samples that cannot be classified correctly are rejected. 

However, the purpose now shifts towards learning the newly introduced classes instead of replaying for old classes. Thus, we developed a sampling filter strategy to discard samples confused by multiple classes, and this strategy is grounded in the assumption that the ground truth feature space for each class is compact and distinguishable, which we have reason to believe based on the earlier observation. This process involves utilizing Mahalanobis Distance $D_M$ to measure the distance between each sampled feature and the feature distribution of other classes. Samples are rejected if the distance falls below a threshold $\beta$. After that, we reconstruct the calibrated feature space from the filtered synthetic features, allowing us to sample new and more robust synthetic features, as described in Alg.~\ref{algo:Augmentation}. 
\begin{figure}[t!]
\centering
\includegraphics[width=0.4\textwidth]{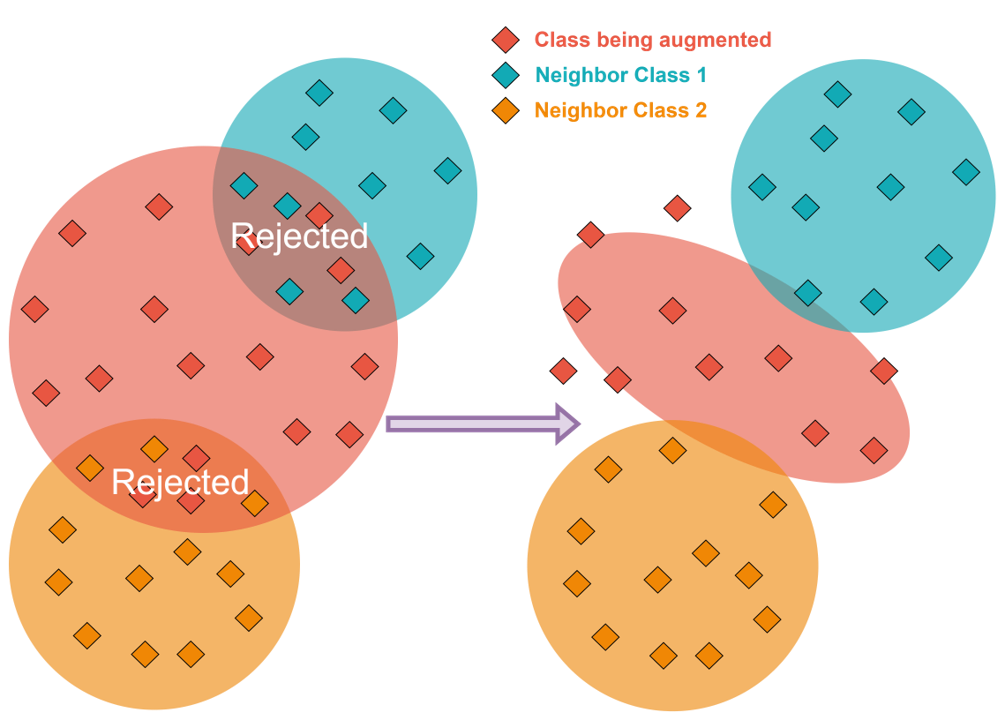}
\vspace{-2ex}\caption{Illustration of sampling process for augmentation. (1) Naively sample enough features from class prototype obtained from real features(limited in few shot setting). (2) Reconstruct the prototype by rejecting sampled features if they are too close to their neighbor class distribution. (3) Re-sampling features from the calibrated prototype for augmentation.}
  \label{fig:feature aug}
  \vspace{-3ex}
\end{figure}

\begin{equation}
\label{mahalanobis}
D_M = \sqrt{(\mathbf{z} - \mu)^\top \sigma^{-1} (\mathbf{z} - \mu)}.
\end{equation}

We investigate the impact of our proposed feature augmentation for new classes in the 5-shot setting.
As shown in Figure \ref{fig:few_shot}, with our proposed augmentation approach, there is an improvement in both global accuracy and IFM consistent across all incremental sessions.
\vspace{-2ex}
\paragraph{Comparison with Other Approaches.}
We compared our results with BOAT-MI under the full-shot setting and TEEN under the 5-shot setting, as shown in Tables \ref{tab:shrec-2017_fs} and \ref{tab:ego_gesture_fs}. Our 5-shot results already outperform BOAT-MI, even in the full-shot setting, which led us to skip the time-consuming 5-shot experiment with BOAT-MI. We also found our performance to be comparable to TEEN, a method specifically designed for FSCIL. For the Shrec-2017 dataset, our approach significantly outperforms TEEN in both accuracy and IFM, and for Ego-Gesture3D, we achieve a much lower IFM. 

\begin{algorithm}

  \KwIn{Training features $ S_i = (Z_k, Y_k) : Y_k \in 
  \mathcal{C}_i$ ,
  all seen class prototypes $\mathscr{P}_j : Y_j \in \bigcup_{j=0}^{i-1} C_j$  }  
  
  \For{$Y_k \in \mathcal{C}_i$ }{
    Compute the mean $\mu_{k}$ and the covariance $\Sigma_k$ from $Z_k$\;
    Sample synthetic features $\hat{Z}_k$ from $\mu_{k}$ and $\Sigma_{k}$ using Eq. \eqref{sampling} \;
    Filtered synthetic features $\hat{Z}_k \newline = \{ \hat{z}_k \mid \hat{z}_k \in \hat{Z}_k, \; D_M(\hat{z}_k, \mathscr{P}_j) < \beta \}$;
     \textcolor{gray}{Here,  $\mathscr{P}_j : j \neq i , Y_j \in \bigcup_{j=0}^{i} C_j$ is the class distribution for all other classes} \;
    Re-compute the mean $\hat{\mu}_k$  and the covariance $\hat{\Sigma}_k$\;
    Re-sample synthetic features  $\hat{Z}_k$ from the calibrated prototype\;
  $Z_k = [Z_k, \hat{Z}_k]$ \textcolor{gray}{\# combine fake features and real features} \;
  }
    \caption{Synthetic Feature Augmentation}
  \label{algo:Augmentation}
 \end{algorithm}

\section{Experiments}
\label{sec:Experiments}
\paragraph{Dataset.}
Following the experimental protocol established by BOAT-MI\cite{boat-mi}, we evaluated our approaches on two hand gesture skeleton datasets, \textit{Shrec-2017} \cite{shrec} and \textit{Ego-Gesture3D} \cite{egogesture,boat-mi}. Furthermore, we further evaluated a new body gesture skeleton dataset, \textit{NTU-RGB+D} \cite{ntu}. (Detailed information can be found in the appendix.)

\vspace{0mm}
\paragraph{Experiment Setup.}
We maintain the incremental task split consistent with BOAT-MI \cite{boat-mi}. For the Shrec-2017 dataset, the base class size is set to 8, with 1 class added incrementally. For the Ego-Gesture3D dataset, there are 59 base classes with 4 new classes added at a time. For DFCIL, we conduct three trials for each dataset with different incremental class orders, following the approach of BOAT-MI \cite{boat-mi}. For FSCIL, we perform three trials with different shot selections but the same incremental class order. Unlike some existing FSCIL methods that run only a single trial, we emphasize the importance of multiple runs. As shown in Table \ref{fig:few_shot}, the standard deviation of the metrics can exceed 15\% in all trials, particularly when the shot size is limited, making the impact of outliers significant. 

\section{Conclusion}%
\label{sec:Conclusion}
DFCIL for gesture recognition has big real-world implications. In this study, we observe skeletons inherently possess a greater generalization ability to new classes and therefore propose a synthetic feature sampling strategy that can replay old class data and augment new class data. Compared to previous approaches, our method not only achieves greater performance, but is also much more efficient, better suited for the gesture recognition domain, which primarily operates on edge devices.

\newpage
{\small
\bibliographystyle{ieee_fullname}
\bibliography{egbib}
}

\end{document}


\title{Supplementary Materials}

\maketitle

\begin{table*}[!]
\small
\centering
\caption{Comparison of DFCIL performance with different size of replay buffer in Shrec-2017  \vspace{-1ex}}
\begin{adjustbox}{width=2\columnwidth,center}
\begin{tabular}{cccccccccccccccc}
\toprule
             &    Task 0 &    \multicolumn{2}{c}{Task 1} &    \multicolumn{2}{c}{Task 2} &    \multicolumn{2}{c}{Task 3} &    \multicolumn{2}{c}{Task 4} &    \multicolumn{2}{c}{Task 5} &    \multicolumn{2}{c}{Task 6} & \multicolumn{2}{c}{Mean (Task $1 \rightarrow 6)$} \\

\midrule
        
          Buffer Size &       & G$\uparrow$ & IFM $\downarrow$ & G$\uparrow$ & IFM$\downarrow$ & G$\uparrow$ & IFM $\downarrow$  & G$\uparrow$ & IFM $\downarrow$  & G$\uparrow$ & IFM $\downarrow$  & G$\uparrow$ & IFM$\downarrow$ & G$\uparrow$ & IFM$\downarrow$ \\ 
          \hline

        10 & 91.9  & 87.1 & 6.0 & 84.0 & 1.3 & 78.8 & 5.1 & 78.9 & 2.5 & 75.0 & 4.6 & 70.8 & 8.5 & 80.9 & 4.0\\ 
        20 & 91.9 & 88.0 & 5.2 & 85.9 & 2.0 & 82.5 & 3.0 & 82.6 & 2.4 & 78.5 & 1.5 & 76.9 & 2.9 & 83.8 & 2.4\\
        50 & 91.9 & 89.8 & 2.9 & 86.9 & 1.6 & 85.3 & 2.7 & 83.9 & 2.7 & 81.8 & 1.6 & 79.8 & 2.1 & 85.6 & 2.0\\
        100 & 91.9  & 90.3 & 3.9 & 87.5 & 1.6 & 86.4 & 2.5 & 84.2 & 2.5 & 82.5 & 1.4 & 81.2 & 1.8 & 86.3 & 2.0 \\
\bottomrule
\end{tabular}
\end{adjustbox}
\label{tab:shrec-2017}
\end{table*}

\begin{table*}[!]
\small
\centering
\vspace{-3mm}
\caption{Results of TEEN for data free class-incremental learning over six tasks in Image datasets. \vspace{-1ex}}
\begin{adjustbox}{width=2\columnwidth,center}
\begin{tabular}{lccccccccccccccc}
\toprule
            Method &    Task 0 &    \multicolumn{2}{c}{Task 1} &    \multicolumn{2}{c}{Task 2} &    \multicolumn{2}{c}{Task 3} &    \multicolumn{2}{c}{Task 4} &    \multicolumn{2}{c}{Task 5} &    \multicolumn{2}{c}{Task 6} & \multicolumn{2}{c}{Mean (Task $1 \rightarrow 6)$} \\

\midrule
        
          &       & G$\uparrow$ & IFM $\downarrow$ & G$\uparrow$ & IFM$\downarrow$ & G$\uparrow$ & IFM $\downarrow$  & G$\uparrow$ & IFM $\downarrow$  & G$\uparrow$ & IFM $\downarrow$  & G$\uparrow$ & IFM$\downarrow$ & G$\uparrow$ & IFM$\downarrow$ \\ 
          \hline

    CIFAR100 \cite{lwf} & 75.5  & 72.2 & 13.5 & 69.1 & 18.0 & 65.9 & 21.2 & 63.4 & 21.6 & 61.0 & 20.4 & 59.5 & 18.7 & 66.7 & 16.2
\\
     CUB200 \cite{lwf} &  79.4 & 75.2 & 7.0 & 73.5 & 9.3 & 69.6 & 14.2 & 70.8 & 9.4 & 68.2 & 9.2 & 69.6 & 7.1 & 72.3 & 8.0
     \\
    MiniImageNet \cite{lwf} & 75.3  & 70.8 & 10.1 & 68.2 & 11.7 & 65.7 & 12.6 & 63.7 & 12.0 & 62.2 & 12.1 & 60.2 & 12.7 & 66.6 & 10.2 \\

\bottomrule \vspace{-5ex}
\end{tabular}
\end{adjustbox}
\label{tab:TEEN}
\end{table*}

\begin{table*}[!]
\small
\centering
\vspace{-1mm}
\caption{ Results (average of 3 runs with different class order) for data free class-incremental learning over six tasks in NTU \vspace{-1ex}}
\begin{adjustbox}{width=2\columnwidth,center}
\begin{tabular}{lccccccccccccccc}
\toprule
            Method &    Task 0 &    \multicolumn{2}{c}{Task 1} &    \multicolumn{2}{c}{Task 2} &    \multicolumn{2}{c}{Task 3} &    \multicolumn{2}{c}{Task 4} &    \multicolumn{2}{c}{Task 5} &    \multicolumn{2}{c}{Task 6} & \multicolumn{2}{c}{Mean (Task $1 \rightarrow 6)$} \\

\midrule
        
          &       & G$\uparrow$ & IFM $\downarrow$ & G$\uparrow$ & IFM$\downarrow$ & G$\uparrow$ & IFM $\downarrow$  & G$\uparrow$ & IFM $\downarrow$  & G$\uparrow$ & IFM $\downarrow$  & G$\uparrow$ & IFM$\downarrow$ & G$\uparrow$ & IFM$\downarrow$ \\ 
          \hline

   TEEN & 88.6 & 84.4 & 15.5 & 81.4 & 15.0 & 78.6 & 14.0 & 76.6 & 13.3 & 72.8 & 15.0 & 70.7 & 14.2 & 79.0 & 12.4
    \\
    SFR-CE & 87.9  & 80.9 & 3.4 & 79.4 & 1.8 & 78.1 & 0.7 & 76.1 & 2.2 & 74.3 & 3.2 & 71.8 & 2.4 & 78.4 & 1.9
    \\
    SFR-CL & 84.3  & 79.9 & 5.1 & 80.0 & 1.0 & 77.8 & 0.6 & 77.0 & 1.5 & 74.4 & 2.9 & 72.9 & 2.7 & 78.0 & 2.0
     \\

\bottomrule \vspace{-5ex}
\end{tabular}
\end{adjustbox}
\label{tab:ntu}
\end{table*}
\section{Experiment Details}
\subsection{Datasets}
We employ three skeleton datasets for experiments, detailed as follows:

\par \noindent \textbf{Shrec-2017} \cite{shrec} comprises of 14 coarse and fine-grained gestures with a training-validation and testing sets of 1980 and 840 samples from a disjoint set of 20 and 8 subjects, respectively. 
\par \noindent \textbf{Ego-Gesture3D} is the 3D skeletal version of the Ego-Gesture\cite{egogesture}. Its original RGB-D dataset contains 14416 training, 4768 validation, and 4977 test samples and there are 83 gesture classes in total. BOAT-MI\cite{boat-mi} constructed its skeletal version through using MediaPipe \cite{mediapipe} to extract 3D skeletons. 

\par \noindent \textbf{NTU-RGB+D}\cite{ntu} is a large human action dataset and comprises of 56880 samples of 60 action classless collected from 40 subjects. It provides multi-modality information including depth images, 3D skeleton, RGB frames, and infrared sequences and we used skeleton modality only.  The task split is defined as a combination of 36 base classes and remaining 24 classes equally divided into 6 incremental sessions. Similarly, three trials with different class orders are evaluated. 
\newline

Image datasets are out of our scope, but we reproduced TEEN\cite{prototype_calibration} on the following three datasets to investigate the overall DFCIL performance difference between image and skeleton datasets. 
Our subset split exactly follows the previous methods \cite{prototype_calibration}.  
\par \noindent \textbf{CIFAR-100}\cite{cifar} compromises of 60000 color images of 100 classes. There are 500 training images and 100 testing images per class. The whole dataset is divided into 60 base classes and the remaining 40 classes are divided into 8 incremental sessions. 

\par \noindent \textbf{CUB-200} \cite{cub} compromises of 11788 color images of 200 classes all belonging to birds and is widely used for fine-grained visual categorization task.  There are 5994 training images and 5794 testing images. The whole dataset is divided into 100 base classes and the remaining 100 classes are divided into 10 incremental sessions. 

\par \noindent \textbf{MiniImageNet} \cite{imagenet} is sampled from the raw ImageNet and compromises of 50000 training images and 10000 testing images, evenly distributed across 100 classes. The whole dataset is divided into 60 base classes and the remaining 40 classes are divided into 8 incremental sessions.

\subsection{Implementation Details}
Following BOAT-MI\cite{boat-mi}, we utilize the DG-STA \cite{dg-sta} architecture as our base model for Shrec-2017 and Ego-Gesture3D with a frame length of 8 and a batch size of 64. The base model is trained with either cross-entropy and supervised contrastive loss, with Adam optimizer (lr = 1e-3, $\beta_1 = 0.9$ and $\beta_2 = 0.999$) for 100 epochs. We utilize a simple linear layer as our classification head, and the linear layer is trained with Adam optimizer (lr = 1e-1, $\beta_1 = 0.9$ and $\beta_2 = 0.999$) for 50 epochs and the best model is selected for evaluation. 

For NTU, we utilize the ST-GCN \cite{st-gcn} as our base model with a frame length of 100 and a batch size of 32. The training detail is the same as above except the base model is trained for 50 epochs instead. 

The synthetic feature size for both replay and augmentation is set to 100 per class with an initial candidate pool size of 300, followed by filtering and then randomly picking 100 as the final feature pool. A dataset can be constructed with all the synthetic features and is replayed/augmented with sample size =  $\alpha$ times the newly introduced data batch size for each iteration in a rolling way. The hyper-parameter $\alpha$ is selected to be 8.

The training details of TEEN on image datasets exactly follows their setup, except all training data are used instead of 5 samples per class.

The threshold $\beta$ for synthetic feature augmentation used in FSCIL is initially set to 30 and the rest set up is the same as in DFCIL. If the target size of sampled feature can't be met after filtering , we can iterativelly reduce the $\beta$ until sufficient features are sampled.
\begin{figure*}[h!]
\centering
\includegraphics[width=\textwidth]{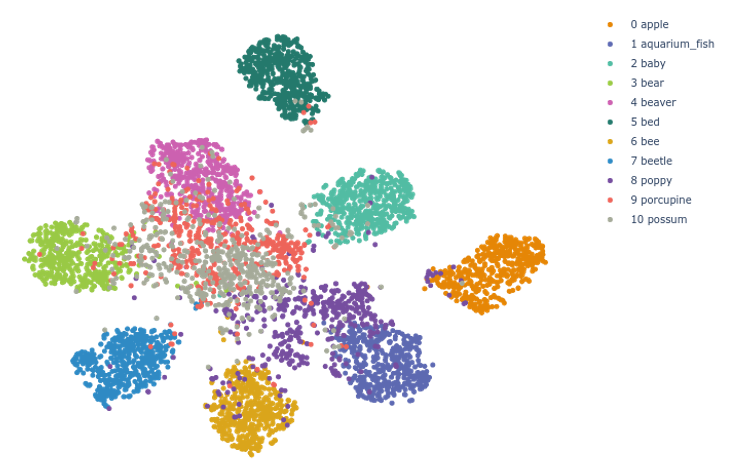}
\vspace{-2ex}
\caption{A toy experiment on image dataset CIFAR100, three new classes were added to a model pre-trained on eight base classes (8,9,10 are three newly introduced classes). Unlike the pattern observed in the skeleton experiment, these newly introduced image classes do not form compact and distinguishable feature spaces. As a result, there is significant overlap between the feature spaces of different classes.}
  \label{fig:image_toy_experiment}
\end{figure*}
\begin{figure*}[h!]
\centering
\includegraphics[width=0.9\textwidth]{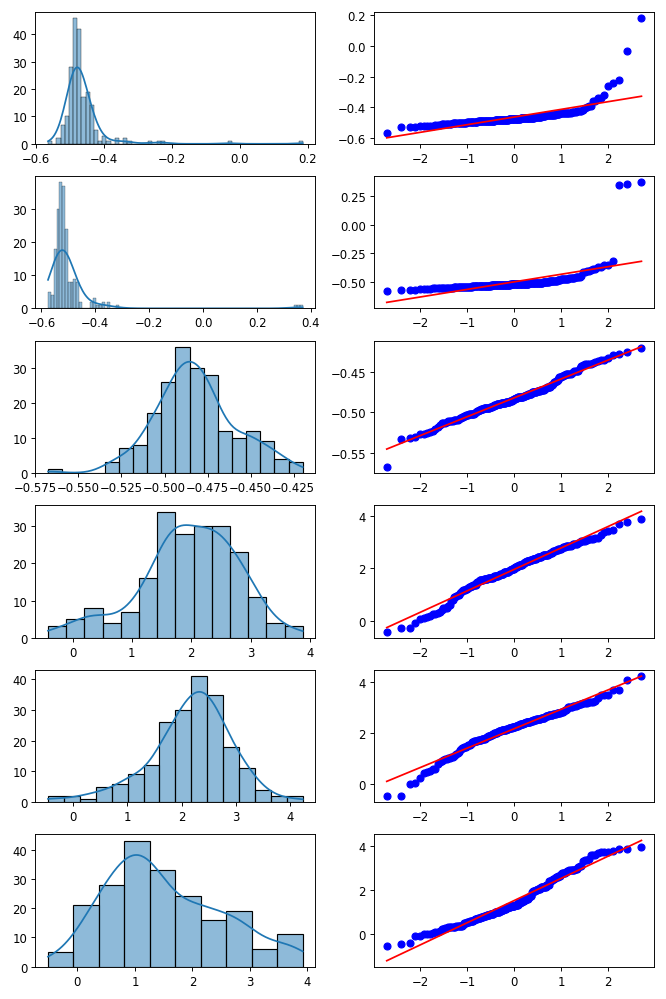}
\vspace{-2ex}\caption{Histograms and Q-Q plots for the top six principal components of the class feature space of a randomly selected class prototype in Shrec-2017, with each row indicates a component.}
  \label{fig:normality_inspect}
  \vspace{-2ex}
\end{figure*}
\section{Discussion}

\paragraph{Performance on body motion dataset} BOAT-MI\cite{boat-mi} only evaluates on two hand gesture datasets, we test our findings and approaches on  a new body motion skeleton dataset - NTU. Since BOAT-MI involves a very time consuming model inversion technique (their implementation of model inversion doesn't support batch processing), it takes too much time when evaluated on the much larger NTU dataset. Therefore, we only compared our approach against TEEN \cite{prototype_calibration}  which requires less computation but has already shown superior performance compared to BOAT-MI. As shown in Table. \ref{tab:ntu}, our approach can also adapt to human body skeleton dataset as well. 

\paragraph{Normal distribution validation}
To verify the correctness and feasibility of the Gaussian assumption in our approach, we plot histograms and quantile-quantile (Q-Q) plots for the top six principal components of the class feature space to  visually inspect the distribution of the feature space as shown in Fig.\ref{fig:normality_inspect}. In most cases, the histograms exhibit strong Gaussian characteristics, and the Q-Q plots largely align with the identity line. This indicates that modeling the class feature space as a Gaussian distribution is reasonable.  However, some samples are concentrated in the tails as well, forming a tailed distribution. For future work, incorporating a prior to ensure that the features of an encoder follow a normal distribution could be a potential solution.
\paragraph{Combine with other approaches}
Our approach doesn't involve any special training or fine-tuning design of the encoder. This allows it to be easily integrated with other methods to further enhance performance. It can be combined with improved training strategies for the base model, such as FACT\cite{forward_compatible}, SAVC\cite{cl_incremental} or fine-tuning techniques for better adaptation to newly introduced classes.

\newpage\
{\small
\bibliographystyle{ieee_fullname}
\bibliography{egbib}
}